# Computer Vision-based Social Distancing Surveillance Solution with Optional Automated Camera Calibration for Large Scale Deployment

**Sreetama Das[1*], Anirban Nag[1*], Dhruba Adhikary[1], Ramswaroop Jeevan Ram[1], Aravind BR[1], Sujit Kumar Ojha[1], Guruprasad M Hegde[2]**

[1] Engineering Data Sciences, [2] Research and Technology Centre, Robert Bosch Engineering and Business Solutions Private Limited, 123 Industrial Layout, Hosur Road, Koramangala, Bangalore - 560095, India

Corresponding author: sreetama.d2005@gmail.com ORCID: 0000-0002-4523-8915, * equal contribution

## Abstract

Social distancing has been suggested as one of the most effective measures to break the chain of viral transmission in the current COVID-19 pandemic. We herein describe a computer vision-based AI-assisted solution to aid compliance with social distancing norms. The solution consists of modules to detect and track people and to identify distance violations. It provides the flexibility to choose between a tool-based mode or an automated mode of camera calibration, making the latter suitable for large-scale deployments. In this paper, we discuss different metrics to assess the risk associated with social distancing violations and how we can differentiate between transient or persistent violations. Our proposed solution performs satisfactorily under different test scenarios, processes video feed at real-time speed as well as addresses data privacy regulations by blurring faces of detected people, making it ideal for deployments.

## 1 Introduction

COVID-19 pandemic has placed a huge burden on the healthcare system. Currently, clinical management primarily includes prevention, diagnosis, and supportive care to hospitalized patients, since many of the proposed therapies are still at different phases of clinical trials or approved only for emergency use. This has set the stage for developing myriad Artificial Intelligence (AI)-based applications to aid healthcare providers [1] including epidemiological modeling, patient triaging [2], and detecting infected subjects from chest X-rays or CT scans, or using cough sounds [3] or abnormal respiratory patterns [4]. AI-assisted strategies to find new or repurposed therapeutic candidates, or understand viral protein structures for drug design are also being explored [5]. Some of these technologies are still in nascent stages and not clinically validated. Hence, the best strategies in the current scenario involve infection prevention and control through social distancing, use of face masks, frequent hand sanitization, and contact tracing [6].

Social distancing involves reducing person-to-person contact by enforcing a minimum physical distance among people in public places, usually 2 meters and generally reducing public gatherings. Such strategies help to reduce the infection rate and delay the onset and size of the epidemic peak [7], and would be necessary for the continued operation of the places of economic activity. However, such requirements are contrary to long-practiced human behavior. Hence, automated and non-intrusive solutions that assist in social distancing compliance can be useful in these times.

Computer vision (CV)-based solutions are particularly well-suited for the automated monitoring of social distancing compliance. Advancements in computer vision algorithms and, notably, in convolutional neural networks (CNN) have led to the development of several object detection algorithms capable of detecting people in video feeds [8]. Similarly, multiple camera calibration techniques have been proposed by researchers to infer distances in real-world coordinates from a 2D image [9],[10] which subsequently help in tracking the number of violations over time.

Technological solutions to monitor social distancing compliance need to satisfy certain criteria to justify the benefits of their implementation. Such a solution should be fast enough to work with real-time video feeds so that timely interventions may be undertaken if the situation demands. The detection algorithm should be robust enough to work accurately under different situations like occlusions occurring in somewhat crowded scenes or detect people irrespective of their poses. Finally, such a system should comply with privacy regulations by not storing or displaying personally identifiable information.

Several CV-based commercial solutions have been developed recently to monitor social distancing [11],[12]. However, not all such solutions provide details of the approach, performance metrics, or the rationale guiding the choice of detection algorithm architecture. Other studies either propose a theoretical approach to a social distancing solution or lack details of considerations for a real-time implementation [13],[14].

This paper describes our approach and field experience while developing and deploying a real-time social distance monitoring solution. We discuss different camera calibration approaches – one requiring manual intervention and the other, fully automated – and their pros and cons during implementation. All steps assume a monocular camera setup, which is cheaper to deploy than stereo cameras. We rationalize our selection of different people detection algorithms, based on benchmarking in literature, and discuss about metrics that will help to understand the extent of compliance for social distancing norms or dissect causes for violation. Finally, we discuss some of the challenges with people detection in real-world situations and propose possible solutions to mitigate the same.

## 2 Application Description

The workflow of our social distance monitoring application is as described in Fig. 1. The algorithm engine consists of five components – people detection, camera calibration, distance estimation, people tracking, and alerts generation, which are described in the following sections. The application is deployed as a hybrid engagement of cloud-based model training and edge infrastructure-based model inferencing.

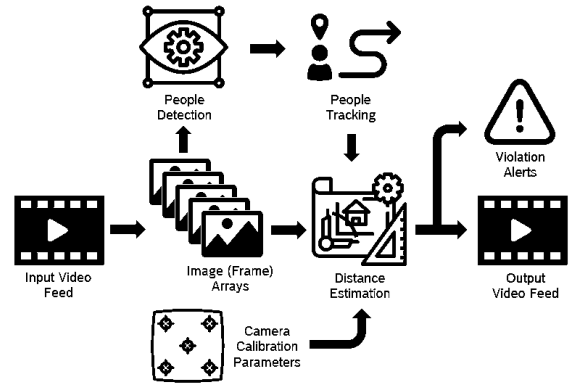

**Fig. 1** Social distancing surveillance application workflow

### 2.1 People Detection

The first step towards monitoring people's movement is to detect people present in a scene and estimate their bounding box coordinates. For this purpose, we utilize the YOLOv3-416 [15] object detector with darknet-53 as the backbone, pre-trained on the MS-COCO dataset [16], and extract the detection results for only the people class. However, our solution is not tightly coupled with any specific framework and can easily be replaced with other state-of-the-art people detector models.

### 2.2 Camera Calibration

The solution, as designed to be deployed for monocular cameras, requires calibration to map image pixel coordinates to real-world coordinates. Our camera calibration module provides flexibility to choose between a tool-based calibration and an automated calibration option.

#### 2.2.1 Tool-based Calibration

In the tool-based calibration method, the user needs to utilize an OpenCV-based tool interface provided as part of our solution to select a rectangular region in the perspective view, all points of which are in the same plane (preferably the ground plane). This region (marked in yellow in Fig. 2) should be clearly visible as a rectangle in the bird's eye view obtained after applying an Inverse Perspective Transformation. A reference length is chosen in the perspective view, which provides the number of pixels corresponding to 2m distance in the bird's eye view. An additional similar reference length can be chosen for the robustness of the calculation.

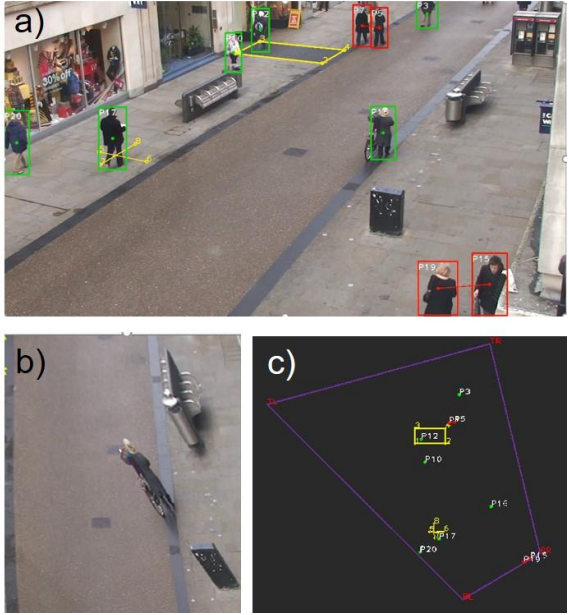

**Fig. 2** Overview of the tool-based calibration. a) Selection of the rectangular region for applying Inverse Perspective Transform. b) Warped top view generated from the Perspective image. c) Bird's eye view showing the rectangular selection, and the locations of the detections.

### 2.2.2 Automated Calibration

Our automated calibration technique doesn't require any tool interface. The heights of the detected bounding boxes (in pixels) are proportional to the projected heights of people from 3D space onto the 2D plane of the camera image. As people keep moving towards or away from the camera, the projected height changes, and it can be expressed as,

$$F = 2 \times \sqrt{1 + 4\frac{xmin+xmax-W}{2H} \times \cos(p) \times \tan\left(\frac{x1}{2}\right)^2} \times \frac{x0}{\sin(x2-p)} \times \cos(p) \times \tan(\frac{x1}{2}) \times \frac{ymax-ymin}{H} \quad (1)$$

where,
p = (0.5 − y*min*/ H) × x1
b = (x*min*, y*min*, x*max*, y*max*) are detection bounding box coordinates,
X = [*x*0, *x*1, *x*2] are camera parameters,
*x*0 = height of the camera from ground,
*x*1 = viewing angle (vertical) of the camera,
*x*2 = inclination angle (vertical tilt) of the camera,
*H*, *W* = frame height and width respectively.

The workflow of this module is as shown in Fig. 3. Similar work by Jung et al. [10] comprises an additional tracker component where each individual is tracked, and the bounding box coordinates of a single person are input to an optimizer to minimize the relative error between randomly sampled pair of estimated heights, whereas Brouwers et al. [17] use comparatively less accurate hog feature-based detectors to separately localize head and feet of pedestrians in order to estimate camera parameters from vanishing point geometry.

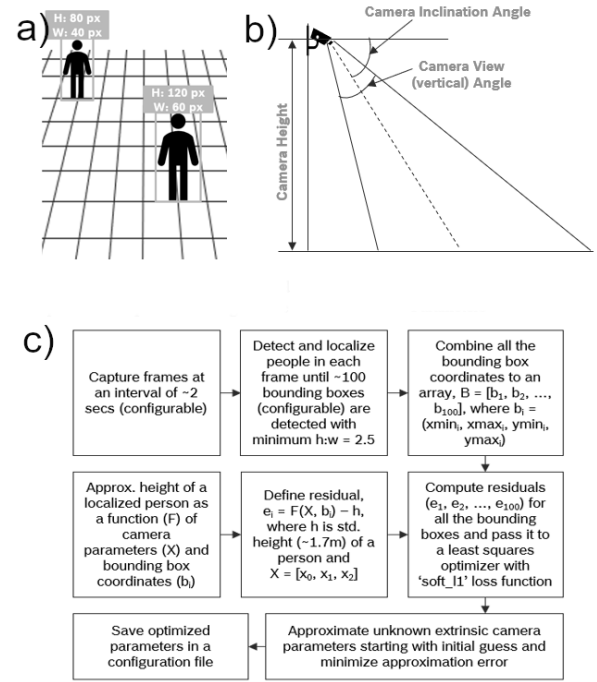

**Fig. 3** Automated camera calibration. a) Changes in size of the detection box, based on the position of a person in the image frame. b) Estimated camera parameters. c) Automated calibration workflow

### 2.3 Distance Estimation

Once the camera calibration parameters and bounding box coordinates of all people visible in an image frame have been obtained, the next step is to estimate the pairwise distances among the people to help in monitoring compliance to social distancing. The mid-point coordinates of the base of each detection box are considered as the location of each person. In case of tool-based calibration, Euclidean distances are calculated among these coordinates in the bird's eye view and compared with the reference distance, to infer whether social distancing is violated.

In the automated method, a circle of radius 1 m is estimated around each person's feet. The projections of these circles in the image plane are ellipses, whose major and minor radii are dependent on the camera parameters and the bounding box coordinates of detected people. Once the position and dimensions of ellipses corresponding to each person are known, we check for overlaps between two or more ellipses. Such overlaps indicate violations of the social distancing policy (Fig. 4). The computations for calculating the ellipse radii and violations for all the detected people are vectorized to increase computation efficiency. In both the approaches, the violations are marked in red, whereas compliances are marked in green.

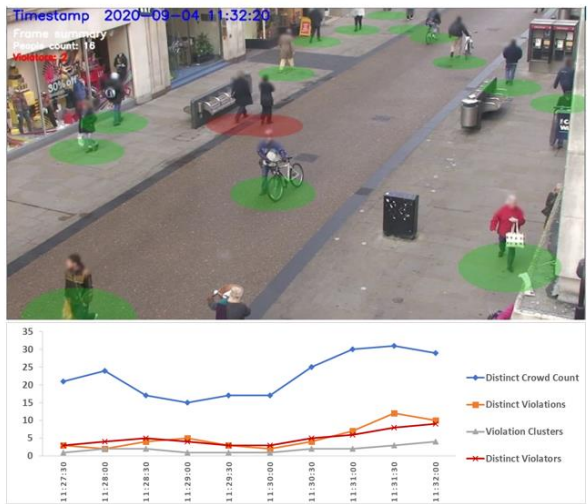

**Fig. 4** Sample application output with automated calibration

### 2.4 People Tracking

The probability of a person getting infected is dependent on the number of distinct violations and the duration of violations [18], which can be computed through frame-to-frame tracking of people. Our solution incorporates motpy, an online multi-object tracker framework which uses IOU of bounding boxes between subsequent frames and Kalman filter to track the detected people present in a scene [19]. Implementation of a tracking algorithm is also expected to mitigate errors in distance calculation due to temporary occlusions, by tracking the preceding detections.

### 2.5 Alerts and Compliance Metrics

In this module, frame-wise violation information over a duration of 30s (configurable) is aggregated to form a violation matrix of size n×n, where, n represents the number of distinct persons detected over all the frames in the aggregation time-window. Each element of the matrix represents the presence or absence of high-risk violations between a pair of tracked individuals. High-risk violations are the instances where the duration ($t_{ij}$) of violation between person i and person j is above a pre-defined threshold (default 5s). Considering pedestrians as nodes and high-risk violations as edges, a graph object is constructed from this high-risk violation matrix (of binary elements), and connected sub-graphs representing violation clusters are identified. The more the linkages within a sub-graph, the higher is the risk of virus transmission. For example, if a node in a cluster is connected to 3 other nodes, it indicates that a person is close to 3 other persons for a considerable duration in that aggregation time-window. The metrics computed as shown in Fig. 5 are plotted for the last 5 minutes and an alert is sent whenever an aggregated metric over this 5-minute duration breaches the user-defined threshold (Fig. 4).

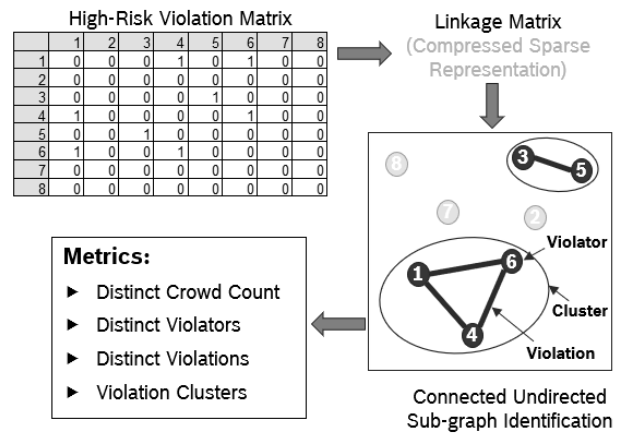

**Fig. 5** Metrics to monitor social distancing compliance

### 2.6 Deployment

To achieve ease of deployment and maintenance, the different components of our application are decoupled into independent modules which communicate among each other via message queues. We also adopted a multi-processing, multi-threading approach to handle multiple IP cameras within the application, thereby enabling scalability. Each video feed is processed by an individual thread on a multi-core edge device, which invokes the algorithm processing unit to process the corresponding image frame and pushes the processed image and associated metrics to a streaming application. The number of camera feeds which can be supported on the edge device during deployment is estimated as (AIP*MAXAL)/SEF, where

AIP: algorithm instance process rate ~ 5 fps.

MAXAL: maximum algorithm instance = minimum of (number of CPU cores, GPU Memory).
SEF: streaming endpoint feed process rate ~ 3fps.
For example, if a system has 12 CPU cores and 16 GB GPU, MAXAL = minimum of (12,16) ~ 12, thereby supporting a total of (5*12)/3 = 20 cameras.

The selection of appropriate video encoding is also important to receive good quality feed with maximum compression. The IP camera encoding allows choosing between multiple formats including MJPEG and H.264. MJPEG is the compilation of separately compressed JPEGs in a sequence, which leads to a high-quality outcome in terms of resolution, with H.264, on the other hand, only some frames are compressed by themselves, while most of them only record changes from the previous frame. This saves a significant amount of bandwidth compared to MJPEG but results in a video of lower quality. As processing is expected to be in real-time, a reduced bandwidth is a key priority to process continuous frames. Also, as the output images are not expected to be of high image quality, H.264 is considered as a preferred encoding.

To address privacy concerns, our application blurs out facial identity in the output video feed displayed. When the detected violations exceed user-defined thresholds of the metrics, messages are announced on a public address system requesting people to segregate. The alerts accumulated over a time interval of 10-14 days may also be analyzed by facility managers or security personnel to take appropriate policy interventions.

## 3 Experiments and Analysis

### 3.1 Object Detection Framework

To select the best-suited object detection model for our solution, we compared the performance metrics of several state-of-the-art models viz. YOLOv3, Faster R-CNN and SSD (Table 1). Available benchmarks on COCO test-dev dataset demonstrate that the mean Average Precision (mAP) is comparable for the FRCNN and YOLOv3 models but much lower for SSD. On the other hand, both SSD and YOLOv3 process images at real-time speed (on a Pascal Titan X GPU) whereas FRCNN is considerably slower [15]. Another important aspect of the people detection problem is to compare how different models perform under small and heavy occlusions. The benchmarking performed on the Euro City Persons test dataset using the log-average miss rate as a performance metric [20] also demonstrates that YOLOv3 and FRCNN perform better than SSD. Considering the trade-off between all these metrics, finally YOLOv3 is chosen as the default object detector of our solution and is deployed without further training. Extensive tests on in-house CCTV footages under different lighting conditions, crowd density, gender, ethnicity, poses of people and occlusions yielded precision and recall values in the ranges 68.8 – 75.6 % and 75.5 – 85.1 % respectively. The best detection performance is observed with the pre-trained YOLOv3 detector for crowd sizes of up to 30 people, provided around 80% of each person was clearly visible.

**Table 1**: Object Detector Benchmarking

| Model | mAP | FPS | Occlusion | |
| --- | --- | --- | --- | --- |
| | | | Small | Heavy |
| YOLOv3-416 | 55.3 | 35 | 17.8 | **37.0** |
| FPN FRCNN | **59.1** | 6 | **16.6** | 52.0 |
| SSD300 | 41.2 | **46** | 20.5 | 42.0 |

### 3.2 Camera Calibration and Distance Estimation

The tool-based bird's eye view calibration process involves manual effort in the selection of the planar points for estimating the inverse perspective transformation, as well as the reference lengths. Such an approach, though relatively more accurate, is not convenient for deployment in enterprises with large numbers of cameras to be monitored. The automated calibration technique is easy to implement on a large scale; however, it occasionally exhibits errors in estimation, especially for people very close to the camera. Hence, it would benefit from restricting the detections to a predetermined region of interest or introducing some correction factors to improve the calculations. The automated calibration approach also expects the people detection module to be highly accurate as it relies on the bounding box information to compute camera parameters. Multiple experiments have been carried out with in-house CCTV footages to identify the ideal range of camera settings, which are reported in Table 2.

**Table 2**: Ideal camera settings for best results

| Ideal camera height | 2.5 – 5.0 meters |
| --- | --- |
| Ideal camera tilt | 5 - 45 degrees |
| Ideal camera-person distance | 10 - 30 meters |

In terms of distance estimation, both the techniques are quite fast, returning outputs in a fraction of a millisecond, thereby ensuring this module doesn't become a bottleneck in real-time processing. Since we are working with video feeds from static cameras, accuracy in camera calibration and of the detection and tracking algorithms will result in good performance of the social distance monitoring application.

### 3.3 Tracking and Compliance Metrics

As our application utilizes tracking-by-detection approach, the performance of the multi-object tracker (MOT) depends on the object detection model in use. MOT metrics obtained on the Oxford Town-center dataset [21] with the YOLOv3 detector and motpy tracker are reported in Table 3. <10% of the tracks are currently lost using a simple tracker like motpy. Moreover, the tracker has an update rate of 260 Hz which is ideal for real-time applications [19].

**Table 3**: Evaluation of Kalman filter-based multi-object tracker

| Precision | Recall | MOTA | MOTP | Mostly Tracked | Partially Tracked | Mostly Lost |
|---|---|---|---|---|---|---|
| 77.6 % | 81.8 % | 57.3 % | 72.1 % | 135 | 77 | 18 |

Finally, we analyzed the variation in our proposed compliance metrics over time on the Town-center dataset. A bulk of the violations are found to be transient in nature, whereas a small proportion are persistent or high-risk violations. Also, we have observed that violations-to-violators ratio is higher for a clustered gathering compared to a queue-like formation and hence raises the risk of transmission.

## 4 Deployment Challenges

Deployment of people detectors for real-time social distance monitoring presents several challenges. Most pre-trained detectors have been trained using the MS-COCO dataset, which contains annotated images of eighty object classes (including people) predominantly in well-lit conditions and from hand-held cameras. As a result, these detectors do not work well on images captured with poorer or non-uniform illumination, substantial shadows, or higher camera heights. Such cases can benefit from re-training the model with annotated video feed captured under the altered conditions and with people class alone. For example, Punn et al. [13] have reported mAP values >0.8 when re-training with people alone, using the YOLOv3 architecture.

All detectors work well in moderately populated conditions in the absence of occlusions, whereas occlusions are very common in crowded public places where the need for social distancing compliance is of critical importance. Tracking algorithms either coupled to the base detector, or advanced one-shot multi-object trackers (for example, FairMOT [22]) help to mitigate temporary occlusions by tracking the preceding detections. An alternate approach in highly crowded scenarios could be the use of density-based trackers [23].

Bounding box-based detection algorithms are also susceptible to changes in people's pose, thereby affecting the location of the base centroid and subsequent distance calculations. Outstretched arms can lead to widened detection boxes, whereas people in a seated posture and partially obstructed by other objects may be completely missed. Pose-based detectors can be a good alternative in such cases [24].

While computer vision-based approaches can identify distance threshold violations continuously on a video feed, not all violations are equally important and will be driven by an understanding of the context. For example, there may be transient violations due to misdetections by the algorithm or flickering in the bounding box dimensions due to pose changes. Such violations can be safely neglected. Additionally, very short (less than 5 sec) violations – which generally occur when people cross each other – may also be discounted. Violations of longer duration need to be monitored, since they indicate people who may have inadvertently come into contact at transit points. Finally, persistent violations by a tracked group of individuals may point to family relations. Our aggregation-based high-risk violation matrix computation approach helps to distinguish the latter case from the former.

Our application utilizes distinct people count, the number of violations and different group sizes to understand adherence to social distancing norms. A time-series analysis of these components would be invaluable to detect trends in time and locations – when and where crowding events tend to occur more frequently. Such information would be immensely helpful in formulating appropriate interventions to mitigate crowding, thereby helping to contain the risk of spreading infection through contact.

A key ethical consideration for vision-based surveillance systems is the individual's right to privacy. Currently, data privacy is addressed in our application by firstly, not storing any data on the system beyond the capture of compliance metrics and secondly, by blurring out faces and not implementing face recognition technology to anonymize personally identifiable information. As a result, the alerts generated when the violations exceed user-defined thresholds are not directed towards any individual. Advanced methods for privacy preservation include the use of generative adversarial networks (GANs) to replace actual faces with synthetically generated facial images [25]. Alternately, if person identification information is used with prior consent, especially at workplaces, persistent violators may be identified and requested to disperse through discreet messages on their personal devices. In addition, multi-camera multi-object tracking [26] may be implemented to determine the number of unique violators at a site, and also identify the people who came into contact with them, thereby supporting contact tracing efforts. However, such measures can only be implemented through clear communication and consent.

## 5 Conclusions

The current pandemic has largely affected physical interactions which involve proximity among people. In this scenario, the implementation of social distancing in public places is an important strategy for ensuring personal health and workplace safety, until vaccines and drugs become available in large volumes for mass use. This work discusses a computer vision-based approach for social distancing surveillance, including an automated camera calibration strategy for easy deployment at scale. We propose the use of time-based thresholds to distinguish between transient and persistent violations of social distancing policy and use metrics like violation clusters to assess risk. We have deployed our solution and achieved real-time performance with satisfactory results under different lighting, crowding, and occlusion. Future experiments to improve our approach include the use of one-shot trackers or density-based trackers, re-training the model with annotated video feeds from the deployment region as well as testing the performance with newer detection algorithms like EfficientDet [27] and YOLOv4 [28], which became available during the course of our application development. Besides, multi-camera multi-object tracking may be implemented to resolve detections across multiple cameras.

## 6 Acknowledgement

The authors thank their colleague Mr. Sairam Yeturi for critical feedback, and the opportunity to learn from the challenges faced during real-time implementation of the methodology.